# *Understanding biology in the age of artificial intelligence*

Elsa Lawrence[1], Adham El-Shazly[+,2], Srijit Seal[+,3,4], Chaitanya K Joshi[5], Pietro Liò[5], Shantanu Singh[4], Andreas Bender[3], Pietro Sormanni[3], Matthew Greenig[3]*

1 Department of Pharmacology, University of Cambridge, UK

2 Department of Philosophy, University of Cambridge, UK

3 Department of Chemistry, University of Cambridge, UK

4 Broad Institute of MIT and Harvard, Cambridge, MA, US

5 Department of Computer Science and Technology, University of Cambridge, UK

Corresponding Author

*mg989@cam.ac.uk

[+]These authors contributed equally.

## Abstract

Modern life sciences research is increasingly relying on artificial intelligence (AI) approaches to model biological systems, primarily centered around the use of machine learning (ML) models. Although ML is undeniably useful for identifying patterns in large, complex data sets, its widespread application in biological sciences represents a significant deviation from traditional methods of scientific inquiry. As such, the interplay between these models and scientific understanding in biology is a topic with important implications for the future of scientific research, yet it is a subject that has received little attention. Here, we draw from an epistemological toolkit to contextualize recent applications of ML in biological sciences under modern philosophical theories of understanding, identifying general principles that can guide the design and application of ML systems to model biological phenomena and advance scientific knowledge. We propose that conceptions of scientific understanding as information compression, qualitative intelligibility, and dependency relation modelling provide a useful framework for interpreting ML-mediated understanding of biological systems. Through a detailed analysis of two key application areas of ML in modern biological research – protein structure prediction and single cell RNA-sequencing – we explore how these features have thus far enabled ML systems to advance scientific understanding of their target phenomena, how they may guide the development of future ML models, and the key obstacles that remain in preventing ML from achieving its potential as a tool for biological discovery. Consideration of the epistemological features of ML applications in biology will improve the prospects of these methods to solve important problems and advance scientific understanding of living systems.

# Introduction

It has become increasingly recognised that artificial intelligence (AI) – and in particular, machine learning (ML) – are accelerating research across a range of scientific disciplines [1], providing a means of identifying patterns in data sets of scale and complexity that preclude their analysis with traditional scientific methods alone. A variety of experimental modalities in biology are now routinely generating data sets that are being combined with ML approaches for analysis, interpretation, and prediction [2-4], defining an important role for these models in the advancement of scientific knowledge and understanding. Our present work is motivated by the observation that many of these approaches are difficult to characterise epistemologically – at least under traditional theories of scientific understanding – and therefore that the philosophical justification for various aspects of ML model design and application in biological research remains underdeveloped. The purposes of this work are thus two-fold: first, to contextualize the use of ML for biological research using modern epistemological conceptions of scientific understanding, and second, to use these ideas to outline key considerations guiding the effective development and deployment of ML models for biological discovery.

## Conventional Notions of Scientific Understanding

Historically, understanding of natural phenomena has been interpreted as a consequence of scientific explanation [5, 6]. An influential view in this domain has been the Deductive-Nomological Model [7], according to which scientific explanation has a law-like deductive structure in which observed phenomena P are explained by reference to more general principles. Typically, this structure includes features of P to be explained, a "law" that explains these features, and sometimes a more comprehensive "theory" that relates various laws to each other under a single framework. For example, if one seeks to explain the warped appearance of an underwater object (Figure 1, panel A), then one may reference Snell's law, which stipulates how light behaves under refraction, and the theory of classical electromagnetism, which can be used to derive Snell's law from Maxwell's equations of light [8]. By subsuming P under these law-like regularities, explanations for P can be obtained deductively. Importantly, this traditional model also comes with an important methodological assumption about natural phenomena: that they involve law-like regularities that are comprehensible by humans. In more recent literature, some philosophers still maintain that having a correct explanation is both necessary and sufficient for understanding [9, 10]. The emphasis on deductive explanation for scientific understanding neatly maps onto how scientific explanation typically proceeds in physics, which provides many impressive examples of explanatory Deductive-Nomological models, including the theories of quantum mechanics, classical electromagnetism, and general relativity. In terms of predictive power (agreement with physical experiments), these theories have been extremely successful. They are also formulated with abstract mathematical language, capable of describing an immense variety of physical systems and configurations with concise notation.

**Constraints on Understanding in Biology**

The situation is rather different in biological science. The most obvious difference compared to physics is that scientific models in biology are typically formulated in terms of *qualitative* descriptions of the underlying phenomena, which imposes limits on their precision and predictive power [11]. The exact reasons for this difference in approach are difficult to unravel and are to some extent tautological; "physics" is simply the word we use to describe science formulated using abstract mathematical language, and indeed "biophysics" is a discipline involving the study of biological phenomena but is often considered to be a subfield of physics rather than biology [12]. Nevertheless, a concept with interesting implications for the study of biological systems, proposed originally by Stephen Wolfram [13] is *computational irreducibility*, a property which describes processes whose outcome cannot be known unless the process itself is explicitly simulated. Wolfram notes that a key objective of physics is to develop computational "shortcuts" (closed-form expressions) for predicting the behavior of natural systems: calculations that can be performed to reveal the outcome of a process without explicitly simulating each step in its evolution, thereby computationally *reducing* the process. The extent to which such opportunities for reduction exist in natural systems is not well understood, but Wolfram notes that even processes governed by simple laws can give rise to behavior that is provably computationally irreducible. Thus, the intrinsic property of computational irreducibility may explain why deductive approaches are less suitable for modelling living systems. More specifically, we identify three key features of biological phenomena that make them challenging to "reduce" computationally:

1. **Multidimensionality.** Biological systems consist of a huge number of components and multiple scales of organization, with complexity at every level of organization (molecular, cellular, organismal, etc.). A model that effectively describes a particular component of a system may not make accurate predictions about a related phenomena; for instance, a biomechanical model of heart function may effectively describe the dynamics of blood flow, but fail to accurately model the effects of heart disease because it does not account for changes in gene expression.

2. **Conditionality.** Evolution has produced biological systems that change their behavior - sometimes drastically - under changes in their environment. Hence, the conditions under which we *study* biological phenomena may not correctly recapitulate the conditions in which we hope to *understand* those phenomena. For example, *in vitro* bioactivity of drugs cannot be directly extrapolated to *in vivo* effects in humans or animals (Figure 1, panel B, middle). More generally, under our current knowledge, biology in many cases does not carry fundamental and general assumptions that license deductive conclusions in a Deductive-Nomological model, at least to the same extent as fundamental physical phenomena.

3. **Emergence.** Biological systems exhibit properties that are difficult to predict from knowledge of their constituent elements alone. For instance, though the chemical features of the individual amino acids that comprise a protein may be known, it has proven challenging to use that information to predict the three-dimensional conformation assumed by the protein molecule (Figure 1, panel B, left). Though this concept is related to computational irreducibility, we draw a distinction by defining emergence as a property specific to an observer of a system; a property can lose its quality of emergence if the observer discovers new features of the constituent elements that allow them to predict the property that emerges from their interaction. On the other hand, properties of computationally irreducible processes *cannot* be predicted; hence, properties of computationally irreducible processes are a subset of emergent properties.

The conjunction of these features affects the direction of scientific explanation; in biological research, answering scientific questions relies almost entirely on inductive reasoning, where data collected from an experiment or observational study are used to make an inference about a more general phenomenon, like a disease or process. This reliance on empirical evidence to reason about biological phenomena has notable disadvantages, primarily related to the fact that experiments are time-consuming, expensive, and are difficult to conduct in conditions that recapitulate features of real biological systems. Yet on the other hand, due to the multidimensionality, conditionality, and emergence exhibited by biological systems, the formulation of reductive, abstract theories seems unlikely to provide models of biology that are sufficiently predictive to be useful in the real world.

We propose that the need to model intractably complex systems with few law-like regularities motivates and justifies the use of machine learning (ML) models in representing biological phenomena. In this review, we explore the following questions:

- How can ML systems be designed to effectively model biological phenomena?
- To what extent do ML systems themselves understand biological phenomena?
- What are the advantages and limitations of ML systems in mediating human understanding of biological phenomena?

With these questions in mind, we survey recent developments in ML technologies and focus on two case studies in biology: protein structure prediction and single-cell RNA sequencing. Importantly, we do not address the widely explored topic of 'explainable AI' [14, 15]: the question of how AI systems can be designed to facilitate human understanding of their inner mechanics. Instead, our primary focus is on how AI models can be designed and interpreted to aid understanding of biological phenomena. While traditional theories of scientific understanding require an explanation of the target P, various modern epistemological theories have explored aspects of understanding that do not necessarily involve deductive-nomological explanations. Many popular ML approaches used in biological research provide real forms of understanding in these ways, but their contribution does not fit neatly into the traditional picture of scientific explanation.

## Understanding beyond explanation

Recent literature in epistemology of science has explored multiple avenues for conceptualizing scientific understanding beyond explanatory requirements. The motivation for these developments is two-fold. First, as we have explored, many natural phenomena - especially biological systems - exhibit a complexity that resists explanatory treatment and unification, at least with current methods and knowledge. Second, a full account of scientific understanding should not only focus on the outcomes of scientific inquiry - which might be explanatory - but also the methodologies and processes that precede the outcomes, which may not be. In this section we discuss three important aspects of scientific understanding independent of providing explanations.

### *Information compression*

Daniel Wilkenfeld recently proposed an account of understanding in terms of information compression [16] (Figure 1, panel C, left). Given our cognitive limitations, understanding plays an important role in highlighting certain information and backgrounding other information. For instance, handling a long sequence of data or information in its uncompressed form might hinder one from identifying regularities in the data, while encoding it efficiently can enable its effective use and manipulation, a problem that has been explored thoroughly in the Noisy Channel Coding Problem from the field of information theory [17]. In other words, an important aspect of understanding is how one encodes information in a way that exposes its regularities. Formally, Wilkenfeld argues that thinkers with better understanding of a target P can generate more information related to P from a more minimal amount of stored (memorised) information [16]. In other words, for a given amount of information stored by an agent, the agent's understanding is a function of how much additional information they can generate from that "kernel" of data. Likewise, for a given amount of information to be generated, an agent's understanding can be measured by how much information they must retain in order to be able to generate that knowledge.

We posit that this interpretation of understanding is not only relevant to humans, but to AI systems more generally. In particular, ample evidence is emerging that deep neural networks benefit significantly from architectural modifications that allow for the selective discarding of information in their internal representations, often achieved by designing operations that respect inherent symmetries of the data - transformations that conserve important properties of the system(s) of interest [18]. The information compression properties of these systems have also allowed scientists to manipulate their internal representations to facilitate their own understanding, for instance using data visualization methods [19].

*Qualitative Intelligibility*

Another recent development in epistemological theory of understanding has been contributed by Henk de Regt, who argues that one hallmark of understanding is the ability to make accurate qualitative predictions in the absence of precise calculations [20] (Figure 1, panel C, middle). Core to his account is the notion that analytical tools such as data visualization contribute to understanding a target P independent of the explanations they might provide. Through an analysis of case studies in physics, de Regt argues that core to understanding is 'intelligibility'; specifically, he posits that a scientific theory is understood (intelligible) by scientists in a given context if they can qualitatively reason about its consequences without performing exact calculations. Crucially, this account of understanding incorporates the epistemic effects of the tools and methods of science—as opposed to solely the outcome of scientific inquiry (explanations)—into a fuller account of understanding, complementary to Wilkenfeld's theory of understanding as information compression. For instance, information compression can be achieved by data visualization methods can make a model more qualitatively intelligible. This generates several features central to scientific understanding, including the ability to manipulate a model and to predict how the target would behave in counterfactual scenarios.

*Dependency relation modelling*

That scientific understanding is not wedded to explanation has also been emphasized by a recent account proposed by Finnur Dellsén [21] (Figure 1, panel C, right). In particular, he proposes that a key aspect of understanding a phenomenon is to generate a model of that phenomenon's dependence relations. In general, scientific models embed information how different parts of a system correspond and relate to each other. Dellsén argues that models provide understanding by highlighting salient dependency relations in a given context, and this is the aspect of the model of P that a scientist must grasp in order to scientifically understand P. According to Dellsén, a dependency model better represents P to the extent that it effectively depicts the network of dependence that P stands in, measured by both accuracy and comprehensiveness. In general terms, comprehensiveness refers to the breadth of the set of dependencies included in the model, while accuracy refers to the correctness of the relationships elaborated by the model, describing the extent to which they model true causal dependencies in the real world. Dellsén argues that accuracy and comprehensiveness can sometimes come into conflict, in that increasing a model's comprehensiveness may produce a less accurate representation of the true causal effects underlying the target P. Conversely, to increase accuracy one may have to sacrifice a model's comprehensiveness, a process which he compares to the general practice of abstraction. These somewhat competing aspects of understanding explain variations in modelling aims and practices in different contexts.

In illustrating his account, Dellsén provides an analogy with causal graphs which describe causal relationships between elements of the target P, and in doing so highlight and background different pieces of information. Although causal models are desirable in many scientific contexts, Dellsén's account is not necessarily limited to dependencies that are directly causal. Importantly, we note that in complex biological systems the nature of causality is often ill-defined, with phenomena at multiple scales of organization (molecules, cells, tissues, and organisms) meeting criteria for causality. The exact relationship between causality and science controversial topic that we largely omit from the present work, choosing to focus instead on phenomena which meet criteria for *sufficient* and *necessary* causality in biological experiments; that is, phenomena that occur under certain experimental conditions, and do not occur when those conditions are not present. These criteria are required for a theory of biological phenomena to be experimentally validated under the falsification principle [22, 23], and they therefore serve as a suitable starting point for discussing how such phenomena might be scientifically understood.

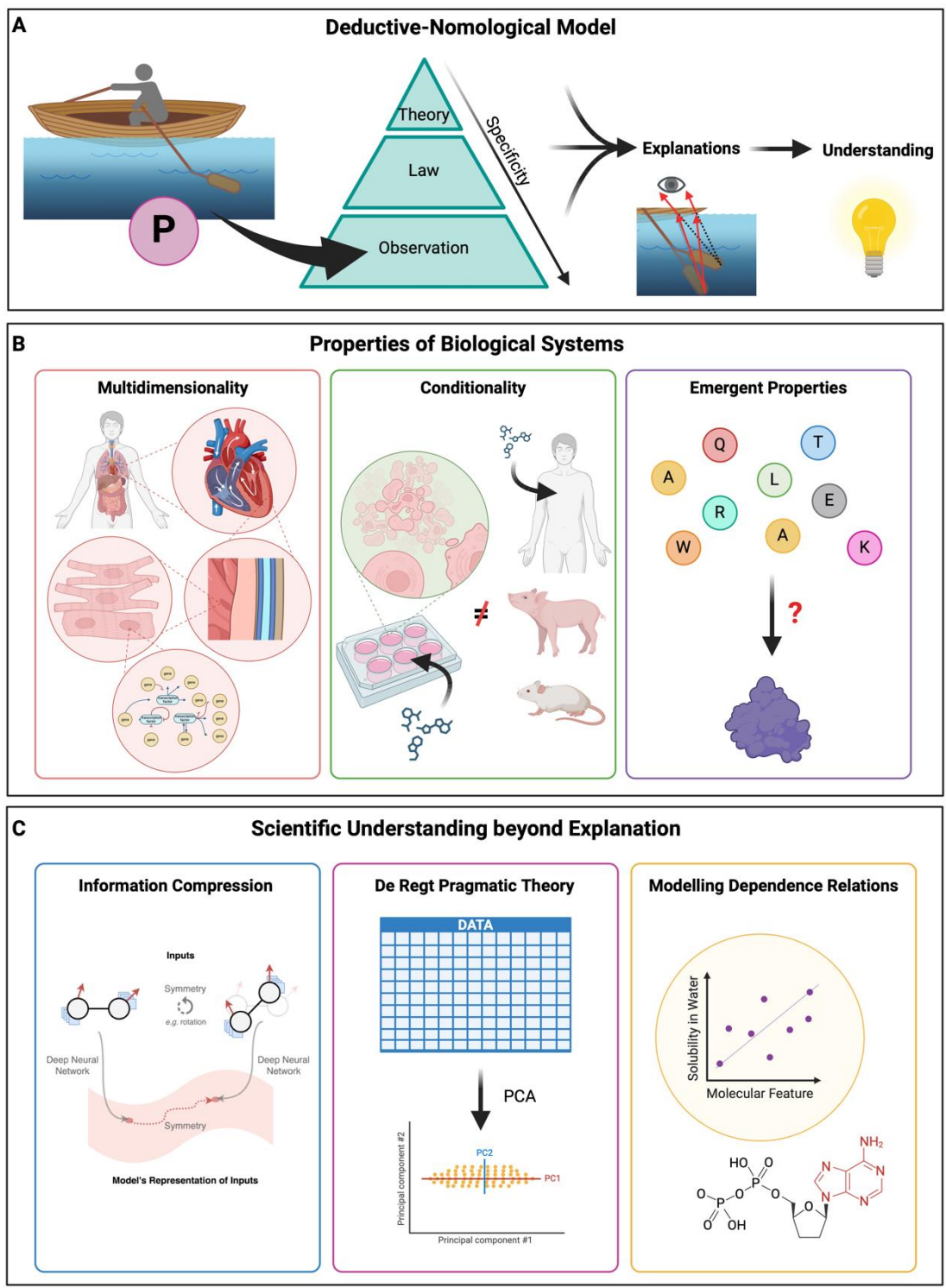

*Figure 1: Understanding biological phenomena is difficult under conventional methods of scientific inquiry. A: Under the deductive-nomological model, scientific explanation has a law-like deductive structure in which observed phenomena P are explained by reference to more general principles. In this case, the observation of the warped appearance of a boat oar can first be explained by Snell's law, which can be derived from the theory of classical electromagnetism. Theories stipulate law-like principles which describe specific observations. B: Three key features of biological phenomena that make them challenging to understand using deductive-nomological explanations: multidimensionality, conditionality, and emergent properties. C: Three epistemological notions of understanding beyond explanation, portrayed in the context of ML: information compression, De Regt pragmatic theory, and modelling dependence relations.*

# Machine Learning in Biology

## Learning from Biological Data

The primary goals of machine learning (ML) are to (1) identify patterns from observations, and (2) generalize that knowledge to new data. To achieve these goals, parametric models are used to mathematically describe patterns or relationships within data and learning algorithms are used to parameterize ("train") those models according to some objective, called an *objective function*, which attempts to measure the model's performance. The objective function typically measures the model's agreement with available data; hence, ML approaches are fundamentally inductive rather than deductive, relying on a finite set of specific observations to formulate models of phenomena that can be applied to arbitrary new data. Importantly, these approaches do not require the same law-like premises necessary for reasoning deductively; instead, a model can be evaluated solely based on its agreement with a set of observations. A model's performance on new data is referred to as generalization performance, and although other measures of a model's performance are important, such as computational cost, interpretability, and fairness, generalization performance is generally considered the most important. Hence, a key concern is *overfitting*, a term used to describe the phenomenon of a model's performance on unseen data being significantly worse than its performance on the training data, indicating that the learning process has captured patterns in the training set that are not characteristic of other data [24]. Overfitting can be driven by a number of factors, but they can broadly be divided into two categories [25]: model-related and data-related. Model-related design choices include the ML model architecture chosen for the task, training procedures, and the form of the objective function, while data-related factors relate to the composition of the data set used for training. The latter deserve particular attention in the context of biological research. Specifically, the properties of multidimensionality and conditionality described previously and in other work [26a, 27b] pose significant challenges in modelling of biological systems with ML.

The multidimensional complexity of biological systems both makes it challenging to collect data that characterize the system of interest in sufficient detail and poses modelling challenges even when such data are available, e.g. in "-omics" analyses [28]. Key challenges in high-dimensional data analysis are described by the curse of dimensionality [29], a term that references the fact that the volume of space increases exponentially as its number of dimensions grows; hence, ML models require exponentially more data to accurately approximate a target function as the dimensionality of the input space increases. These difficulties introduce a tradeoff between the breadth of the input data and the performance of the model. Furthermore, high-dimensional models may be less conducive to scientific understanding by virtue of failing to compress information effectively [16], enable qualitative interpretation [20], or accurately model true causal dependencies [21]. The feature of conditionality also introduces a variety of challenges in ML-driven modelling of biological systems. Fundamentally, ML approaches applied in biology are almost always aimed at modelling *in vivo* phenomena; however, available data are often limited to those collected from experiments *in vitro* or using model organisms. This difference not only makes it difficult to fit models that perform well in their intended setting; it can also introduce a discrepancy between a model's evaluation - usually done using a subset of the available data - and the model's actual performance in the context in which it will be applied. Multidimensionality and conditionality also pose significant challenges in appropriately *labelling* biological data; determining what quantities a supervised ML model should be trained to predict to be useful for its intended purpose. Many biological phenomena of interest are difficult to describe numerically due to their multidimensional nature (e.g. disease states) or significant dependence on other factors (e.g. biochemical properties like water solubility), and it may not always be possible to collect data or parameterise an objective function that accurately represents what the model is intended to achieve. All of these factors combine to make data representation an important design choice in ML applications for biological science.

**Data representations**

In the context of ML, a representation refers to the particular scheme used to describe an object numerically [30]. The representations used to describe data often derive from manual curation of numerical features believed to be relevant to the ML task. However, in biological systems with a huge number of components and few general assumptions, manual selection of features is often infeasible and representations are instead constructed with automated approaches. Principal Components Analysis (PCA) is one such approach popularly applied for "dimensionality reduction", the task of identifying useful low-dimensional representations of high-dimensional data [31]. PCA has become a core tool in unsupervised learning and data analysis more generally, allowing researchers to describe data sets using a small number of "meta-variables" (components) that faithfully represent relationships between the individual data points. If a reasonable amount of variation in the data set can be described using only two or three principal components, the positions of data points along these components can be visualized in a scatter plot, which is useful in many contexts for facilitating qualitative understanding of high-dimensional data [20].

In certain high-dimensional applications, however, PCA's linearity constraint - that components must be described as a weighted sum of input variables - impedes its usefulness. Indeed it is quite common in data sets with large numbers of variables – for instance, -omics datasets that simultaneously measure thousands of molecules – that the top 2 or 3 principal components account for a small proportion of the data's total variation across all variables, which produces a low-dimensional representation that does not accurately represent the high-dimensional relationships between data points [32]. These limitations have prompted the development of non-linear dimensionality reduction techniques, including autoencoders [33], t-SNE [34], and UMAP [35], which eschew the linearity constraint and aim solely to produce a useful low-dimensional representation, typically with the objective of effective data visualization and interpretation. A complementary approach is clustering, which can be interpreted as a data representation method that identifies a discrete representation of the data, consisting of some finite number of categories. Beyond visualisation and interpretability, a vast body of modern research has also explored the representational capabilities of deep neural networks, ML architectures that perform learnable transformations of data to produce internal numerical representations that facilitate the task they are trained to perform [18, 36]. These models have achieved striking success in a variety of tasks in computational biology, including protein structure prediction [3], drug discovery [37], and novel protein design [38]. These innovations and others in ML more broadly have been catalysed not only by improvements in data quality and availability but also by advances in model design. In the following section we describe a key unifying framework of ML system design: inductive biases.

**Inductive Biases: Designing ML architectures**

Inductive biases are the underlying assumptions that an ML model implicitly relies on in order to make predictions on unseen data; since unknown quantities can take a range of possible values, an ML model must make *a priori* assumptions in order to generalise beyond its training set [39]. Inductive biases may be expressed at different stages in a machine learning pipeline, but here we focus primarily on those that are expressed via the model's architectural design choices: the mathematical operations it performs to transform its input into its output, which, in the most general terms, express the inductive bias that the output variables are related to the input variables by the class of function parameterised by the model. While ML models can learn the parameter *values* that dictate the relationship between input and output variables, the *form* of the model - the operations that are used to produce the output from the parameters and input(s) - is typically determined before training begins. A key role of machine learning practitioners is therefore to develop systems with inductive biases that are synergistic to the problem, reflecting high-confidence assumptions typically informed by human intuition or domain expertise. This enables the model to learn patterns important for generalisation and has been described as a means of reducing the dimensionality of the learning task, reducing the amount of data required to approximate a target function within a given input space [18]. Here we characterize and review three classes of inductive bias that are particularly relevant in biology: locality-related, distributional, and symmetry-related.

We define locality-related inductive biases as those which assume that proximity between input entities provides information about characteristics of those entities which may not be positional in nature (Figure 2, panel A). Locality-based inductive biases encode the notion that proximity – whether in a sequence, network, or another space - determines the extent to which entities interact, positing an underlying dependency structure of the target function that relates to the distances between entities [21]. Therefore, ML models with a locality-related inductive bias often enforce constraints on the proximities at which entities in the input data may influence each other in the ML model. A classic example of this inductive bias in practice can be observed in the K-means clustering algorithm [40], which uses euclidean distance between the feature vectors of different instances to define proximity, and identifies similar groups of data points by assigning them to the cluster to which their feature vector has the greatest proximity. This foundational algorithm has found applications in a wide range of biological problems - particularly in analysis of gene expression data - including identification of cell types [41] and inference of gene co-expression modules [42]. Underlying these applications is the inductive bias that observations (e.g. cells or genes) with similar features should be grouped together in an effective clustering. In deep neural networks, the concept of local convolutions - learned functions that operate over small neighborhoods in the input data domain - are now a ubiquitous feature in a variety of layer architectures relevant to biological problems, notable examples of which include protein design models [43], which often enforce constraints on the physical distances at which amino acids can interact, and medical image processing models [44], which typically transform pixels according only to their local neighborhood within the image.

Distributional inductive biases are those which make assumptions related to the random processes that generate data (Figure 2, panel B). These inductive biases are frequently observed in classical statistical models, in which learning algorithms - or "parameter estimation" procedures - are often derived from assumptions about the probability distribution(s) of the variable(s) to be modelled. These assumptions can be conceptualized as a form of information compression in that they allow data sets and populations of arbitrary size to be described using a fixed number of parameters, not only selectively discarding information within the model itself but also facilitating qualitative interpretation and understanding by humans. For instance, the entire class of Bayesian statistical methods is grounded in the principle that existing beliefs about a random process can be expressed using a prior distribution, which describes the likelihood of different parameter values independent of the observed data values [45]. These methods have found a huge range of applications in biological problems ranging from phylogenetic tree inference [46] to gene expression analysis [47], due in part to their ability to express domain-specific knowledge in the form of distributional inductive biases. For example, the widely-used DESeq2 generalized linear model for differential expression analysis of RNA-sequencing data relies on a variety of domain-specific distributional inductive biases, including the assumption of a relationship between the variance (dispersion) and mean expression levels across all genes, which allows their model to more robustly estimate uncertainty in observed gene expression changes [48]. In general, the suite of statistical analysis methods and machine learning models developed for functional genomics data use a wide range of distributional inductive biases, and choosing the most appropriate model for a given data set depends heavily on the conditions under which the data was generated.

The final class of inductive bias we consider is symmetry-related (Figure 2, panel C). The notion of symmetry has received significant attention in machine learning literature, particularly in the emerging field of geometric deep learning, which aims to devise neural network models that respect the invariances and symmetries in data [18]. Symmetries are defined as classes of transformations under which the properties of a system remain consistent in some way, typically formalized in the language of abstract group theory. The core principle of geometric deep learning is that designing models to respect the symmetries of the data - that is, models that behave predictably when symmetry transformations are applied to their input data - improves performance significantly by constraining the learning task. It is believed that the key advantage of encoding symmetry is that it constrains the model to only use information which is relevant to the problem of interest rather than artifactual features of the data representation, an interpretation which aligns well with Wilkenfeld's theory of understanding as information compression [16]. Graph Neural Networks (GNNs) – bespoke neural networks for graph data that incorporate permutation symmetry – best illustrate geometric deep learning concepts for biology [49]. Graphs are a universal language for describing biological systems, ranging from molecules - represented as graphs of atoms - to interactome networks of proteins and maps of synaptic connections in the brain. Importantly, graphs typically have permutation symmetry: the graph structure is not associated with any particular ordering of the nodes. GNNs process graph data by applying a shared set of parameters at each node's local neighborhood and are explicitly structured such that each node's computation is invariant to the permutation of nodes [50], preventing the model from learning information related to node orderings. Interestingly, the transformer [51] – a modern architecture that has powered key deep learning applications in biology [3, 38, 52, 53] – can be described as a GNN operating on a fully connected graph [54]. The key component of the transformer is the multihead attention layer, a permutation-symmetric operation that passes information between all pairs of input objects, with learnable weights determining the interaction between each pair of items. A key insight was that even in domains with permutation *asymmetry*, the transformer can be designed to use ordering information by augmenting the representations of individual elements with numerical features that describe each element's position [51], a technique that has enabled highly successful biological language modelling applications [52, 53], demonstrating how the *absence* of a particular symmetry can also inform model design choices. Other notable examples of symmetry-related inductive biases appear in ML models of three-dimensional biological systems with rotational and translational symmetries, most notably in the context of structural biology, where the global orientation (rotation) and position of a molecule within a coordinate system has no bearing on its biological function. Geometric GNNs and transformers that process molecular graphs equipped with additional geometric features (e.g. bond orientations) have had widespread success in problems such as protein structure prediction, protein design, and protein ligand-docking [3, 38, 55, 56].

The three classes of inductive bias described here - locality-related, distributional, and symmetry-related - appear in a variety of specific forms, but fundamentally all seek to improve the generalization performance of ML systems by encoding scientific expertise and human intuition. While superficially it may appear that these techniques are simply a "trick" to increase ML model performance as measured by the relevant metrics, we propose instead that they represent a transfer of human understanding to ML systems, particularly as characterized by Wilkenfeld's theory of understanding as information compression [16] and Dellsén's theory of understanding as dependency relation modelling [21]. Though the inner workings of ML models can be complex and difficult to interpret, encoding human knowledge in their design enables researchers to obtain robust guarantees about their internal behavior and reason effectively about their properties. The remainder of this work explores key case studies highlighting important applications of ML in modern biological research, highlighting key epistemological principles that underlie the successful design and deployment of ML systems for understanding biological targets.

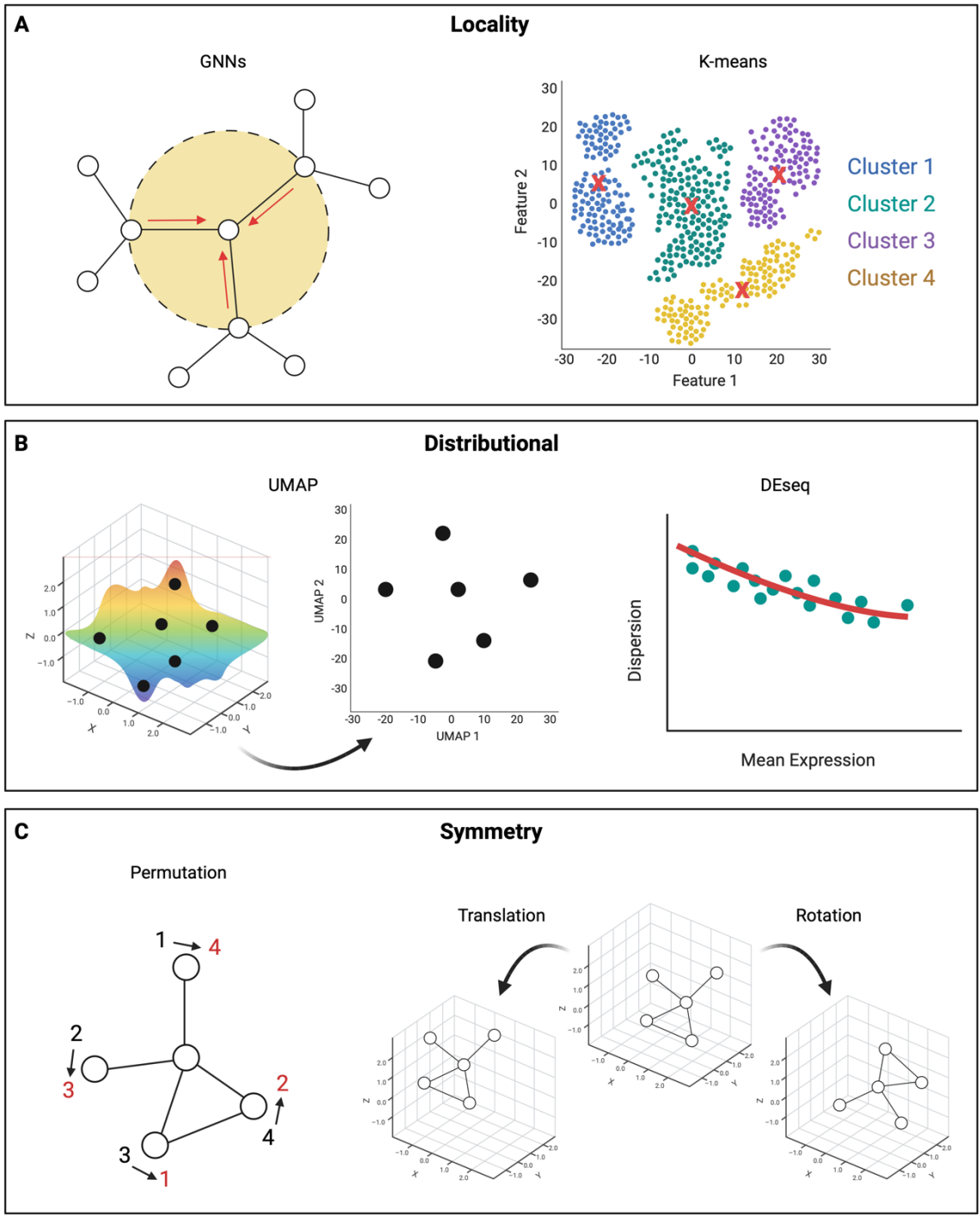

*Figure 2: Inductive biases are a unifying principle across a variety of machine learning approaches to biological problems. We identify three classes of inductive bias commonly used in ML models of biological systems. A: Locality-related: proximity encodes information about relationships between observations. Models with a locality-related inductive bias include GNNs (graph neural networks) and K-means clustering. B: Distributional: the random process that generates observed data has a known form. Models with a distributional inductive bias include UMAP (Uniform manifold approximation projection) and DESeq2. C: Symmetry-related: the property being estimated remains constant under certain transformations of the system. Examples of such transformations include permutation, rotation, and translation.*

# Protein structure prediction

## Introduction

The three-dimensional structures of proteins are a key determinant of their chemical and physiological activity, and application areas ranging from drug discovery to bioprocess engineering benefit from accurate protein structural information. However, structural data is difficult to obtain experimentally, especially compared to the ease with which modern high-throughput sequencing technologies can reveal the sequence of amino acids that comprise a protein of interest. In physical terms, a protein's amino acid sequence specifies only the covalent peptide bonds present in the molecule; it does not directly provide any information regarding the huge number of non-covalent interactions which characterize a typical protein's three-dimensional structure. Yet, strikingly, it was demonstrated in the early 1960s that many proteins can spontaneously refold into their native structures after denaturation, demonstrating that the peptide bond configuration of a protein molecule - that is, its amino sequence - determines the protein's native three-dimensional structure in most cases [57]. These observations have since prompted efforts to develop computational approaches for predicting protein structure from sequence data [58], motivated by the hypothesis that the information needed to encode a protein's three-dimensional shape is contained entirely in its sequence of amino acids.

Though the relationship between a protein's sequence and three-dimensional structure is commonly referred to as the "protein folding problem" [59], it is important to distinguish protein *structure* from protein *folding*. Protein folding is a dynamic process in which a chain of amino acids physically moves and eventually assumes a stable conformational state (in most proteins), while that stable conformational state is the protein's structure; hence, protein folding is the physical process that generates protein structure. Protein folding itself is a field of scientific study that has attracted considerable interest [60], primarily owing to its putative role in neurodegeneration. However, in many cases it is possible to accurately predict the three-dimensional structure assumed by a protein without directly simulating the dynamics of the folding process, showing that, to some extent, the process of protein folding is "computationally reducible" [13]. In this section, we provide a brief review of historical approaches to computational protein structure prediction to introduce the ML paradigm with appropriate context. We then review the key architectural components of AlphaFold2 [3], the most broadly successful AI system for protein structure prediction to date. We conclude by exploring key epistemological questions related to the AlphaFold2 system, including how well the model itself understands protein structure and the extent to which it can advance human scientific understanding of sequence/structure relationships and protein folding.

**History of protein structure prediction**

Historically, approaches to structure prediction have been categorized as either template-free or template-based [58], depending on whether they make use of experimentally resolved "template" structures of other proteins in order to model the structure of the protein to be predicted. Philosophically, these methodological categories can be roughly described as mechanistic (template-free models) and empirical (templated-based models) (Figure 3, panel A). Mechanistic approaches attempt to predict structure by treating the protein and its surrounding environment as a physical system and parameterising the features of the system to reflect our understanding of natural laws. These methods typically rely on a human-designed energy function to calculate the energetic favorability of three-dimensional conformations of the protein molecule. The energy function relates the potential energy of the molecule to its chemical and structural features, and is typically parameterised to maximize agreement with experimental data [61]. Conversely, empirical protein structure prediction methods rely on using known (experimentally-determined) protein structures to infer the structural features of new proteins. These methods are based on the assumption that evolutionarily-related proteins tend to adopt similar structures, which has motivated the development of algorithmic methods for identifying homologous sequences, such as BLAST [62] and later hidden markov models (HMMs) [63]. These methods do not attempt to model how exactly observed patterns of sequence variation relate to the underlying biophysics of protein structure; most modern template-based approaches simply rely on identifying homologous sequences and performing physics-based structural optimisation post-hoc [58]. The modern suite of machine learning-based approaches for protein structure prediction have observed great success integrating these two methodologies [3, 64, 65]; in the mechanistic domain by exploiting physical symmetries to design machine learning models capable of efficiently representing and learning three-dimensional geometric information, and in the empirical domain by training models on large datasets to capture general patterns of sequence variation in proteins.

In recent years a new paradigm of protein structure prediction has emerged, in which end-to-end deep learning architectures are trained to predict folded protein conformations using extensive datasets of experimentally determined structures [3, 52, 64]. In the 14th Annual Critical Assessment of Protein Folding (CASP14), the Alphafold2 system demonstrated astonishing structure prediction accuracy, outperforming competing methods by an order of magnitude, particularly on targets without reliable template structures. In early iterations of CASP, accuracy improvements were primarily attributed to the development of improved statistical methods for identifying sequence homology [62, 63]. By 2016, the organizers of CASP11 had noted that, compared to previous competitions, prediction accuracy had ceased to improve significantly for targets with publicly-available template structures from a close homologue [66]. Some progress, however, was observed in the template-free category, owing to the introduction of novel statistical methods - "evolutionary coupling" techniques - to predict contacts between residues based solely on patterns of sequence covariation between aligned residues in homologous proteins [66]. These techniques were a conceptual precursor to later deep learning models that learn representations of multiple sequence alignment (MSA) data [3, 64, 67]. While single-sequence approaches based on large language models [52, 53] are becoming more competitive [65], MSA-based models - and AlphaFold2 in particular - remain the state of the art for protein structure prediction; hence, we focus solely on AlphaFold2 in this review, noting that many of the general principles we propose apply broadly, across different deep learning-based approaches.

**AlphaFold2 architecture**

In both its original implementation and later in AlphaFold2, the MSA for an input sequence serves as the key piece of information used by the model to predict a protein's structure. AlphaFold2's Evoformer module learns representations of homologous sequences in the MSA, where residue-level features from individual related sequences in the MSA - or clusters of related sequences in large MSAs - are explicitly used as input to the model and processed by the model's attention layers alongside the sequence of interest, to be predicted. Importantly, as a transformer-based architecture, the Evoformer's information compression properties are determined by learnable parameters rather than explicit constraints on the range at which pairs of residues can interact. In the context of protein structure, the benefit of global attention mechanisms that span the entirety of the input sequence are immediately obvious; residues that are distant in the protein's primary structure may still interact in three-dimensional space. We emphasize the observation that these attention mechanisms - in AlphaFold2 and other transformer models - allow the network to discard information while simultaneously providing the opportunity for distant residues to interact, drawing interesting parallels with Wilkenfeld's epistemological theory of understanding as a consequence of information compression [16].

Dellsén's theory of understanding as dependency relation modelling provides yet another framework for interpreting AlphaFold2's understanding of protein folding [21]. Accurately modelling dependencies between residues is at the crux of protein structure prediction, and deep learning models are capable of learning highly informative internal representations of such relationships. However, mapping these internal representations of residue pairs into valid three-dimensional structures is a non-trivial challenge. The initial iteration of AlphaFold directly modelled distances between residues in the input protein [67]; however, these distances were not guaranteed to produce a valid configuration of atoms in three dimensional space, and were instead used solely as an objective in a subsequent torsion angle optimisation, keeping the bond lengths and bond angles within the protein fixed to their chemically-valid values. AlphaFold2 adopts a different approach, centered around two novel layer architectures in the Evoformer known as triangular update and triangular attention modules (triangular update show in Figure 3, panel B). Both layer architectures update the representation of each pair of residues using information from all other pairs of residues involving a residue from the pair of interest. This method is grounded in the triangular inequality in euclidean geometry, which imposes a fundamental constraint on the distance between two points if their distances from a third point are known. However, triangular layers do not impose any explicit constraints on the pair representations learned internally by the network; rather, they only provide a mechanism for modelling dependencies [21] between different residue pairs, parameterising layers to allow them to selectively retain and discard information.

The Structure Module uses the residue and pair representations produced by the Evoformer to generate a three-dimensional configuration of the structure (Figure 3, panel C). The essence of the SM's approach is to model each residue as an independent three-dimensional entity with a rotational component (the orientation of the CA backbone bonds) and a translational component (the coordinate of the alpha carbon), collectively referred to as an orientation frame. Crucially, the orientation frames provide a means of converting between coordinate representations expressed relative to each residue and those expressed within the global coordinate system, using the residue's specific orientation to perform a change of basis. This interconversion between coordinate systems is at the core of the Invariant Point Attention (IPA) layers introduced in Structure Module. IPA calculates affinities between each pair of residues in the protein based on the representations obtained from the Evoformer and the residues' current frames, where the affinity decreases as a function of the distance between two frames, introducing a locality-related inductive bias. Importantly, the use of orientation frames in this context produces an update which is invariant to rigid motions of the entire protein structure; hence, all rotations and translations of the predicted protein structure (which influence the orientation frames) lead to the same update of the underlying embeddings, introducing a symmetry-related inductive bias and providing yet another example of the system's information compression capabilities. The final prediction of the structure module is an atomic-resolution three-dimensional configuration of the protein's amino acids with an estimated confidence score for each residue, trained to recapitulate the likelihood that the residue's predicted conformation matches the ground truth crystal structure.

**Implications for protein science**

How the AlphaFold2 system will shape scientific understanding of protein structure and folding going forward is still uncertain. The complexities of its internal mechanics make the system difficult to directly interrogate and so it remains an open question what it has learned *conceptually* from its training process, a question that has important implications for determining the model's capabilities and potential to assist humans in scientific inquiry. In our summary of the model's architecture we hoped to show that the themes of information compression and dependency relation modelling appear ubiquitously in the system's design, providing a way of conceptualizing the system's understanding of protein structure under Wilkenfeld's and Dellsén's theories of human understanding [16, 21]. While the model's information compression properties are difficult to evaluate empirically, its ability to model relevant dependency relations can be assessed to some extent by evaluating its performance on tasks that are distinct from protein structure prediction but have a similar or related dependency structure. The evaluation results from CASP14 established that the model is capable of producing three-dimensional structures that match experimentally determined crystal structures for a wide range of proteins, robustly showing that the model has strong generalization performance on its training task but failing to provide sufficient evidence that the system is able to reason about biochemistry more generally.

Shortly after its release, foundational work evaluated the AlphaFold2 system on a variety of downstream tasks relevant to the model's general biochemical reasoning capabilities [68]. For instance, an analysis of the distribution of structural elements in the predicted versus experimentally-resolved proteome revealed elements predicted by AlphaFold2 that were not present in the space of experimentally-determined crystal structures (i.e. the training set). One element of particular interest is the beta solenoid, a structural element consisting of repeating beta strand subunits found abundantly in microbial pathogenic proteins and in highly diverse conformations [69]. Experimental characterisation of these structures has not explored the full space of possible conformations, making empirical evaluation difficult but also providing a relevant benchmark to assess AlphaFold2's ability to reason about exotic structural elements. The authors note that the AlphaFold2 models of novel beta solenoids appear chemically plausible, and later work performed a more thorough analysis of the space of predicted beta-solenoid structures and identified a large variety of plausible predicted structures [70]. Another area of interest has been AlphaFold2's ability to predict intrinsically disordered regions (IDRs) of proteins. The potential to use the system's residue structural confidence score (pLDDT) as a predictor of intrinsically disordered regions was first identified by the project expanding the system's coverage to the entire human proteome [71], where pLDDT was shown to be competitive with the state of the art method for prediction of disorder on the critical assessment of intrinsic disorder (CAID) benchmark dataset [72]. Importantly, though IDRs do appear in AlphaFold2's training set, the configurations in which they appear in experimentally-determined crystal structures are likely to be arbitrary; the model's propensity to identify them accurately therefore reveals capabilities for general reasoning about sequence-structure relationships in proteins. More recent work has even indicated that AlphaFold2 has learned information related to protein dynamics [73, 74], showing that, for instance, ensembles of structures generated by adding noise to AlphaFold2's forward pass are able to sample structural states of proteins that are inaccessible via molecular dynamics simulations [74]. These observations provide evidence that, in addition to learning how to perform its training task of predicting crystal structures, the system has accurately modelled some of the underlying dependency relationships that determine a protein's three-dimensional structure.

Though these results indicate that the AlphaFold2 system itself understands protein biochemistry to some extent (at least under Dellsén's account), the model's potential to advance human scientific understanding of protein folding is a separate question. Detractors are likely to point out that, despite all of its capabilities, the model in its current form can only be used as a black box and has so many parameters that it fails to provide any human-accessible insight into the mechanics of the underlying phenomenon. But perhaps there are limits to the extent to which protein folding, as a physical process, is computationally reducible into simpler, lower-dimensional descriptions. Proteins, like any other molecule, are fundamentally governed by physical principles; it is generally accepted that typical proteins are large enough to be modelled as classical mechanical systems [75]. Though classical mechanics is seemingly capable of explaining protein folding using a small number of terms [76], the mathematical formulation can be misleading since choices of notation and conceptual abstraction can mask underlying complexity. In reality, the actual number of computations required to calculate a protein's folded state in a physical simulation is immense, and though progress has been made to identify pockets of computational reducibility within such systems, it still remains infeasible to perform physics-based simulations for protein structures larger than 50 amino acids [77, 78]. AlphaFold2, on the other hand, requires a prolonged training period of weeks but is able to predict structures in seconds or minutes, performing far fewer actual calculations at inference time than are required to simulate a protein with molecular dynamics. We therefore propose that AlphaFold2, despite its complexity, constitutes a "better-reduced" [13] description of protein folding than our best physical theories. Similarly to how progress in theoretical physics continues to invent more useful abstractions over time, developments in how we mathematically describe and conceptualize ML models may one day allow us to better understand the mechanics of AlphaFold2 and protein folding itself.

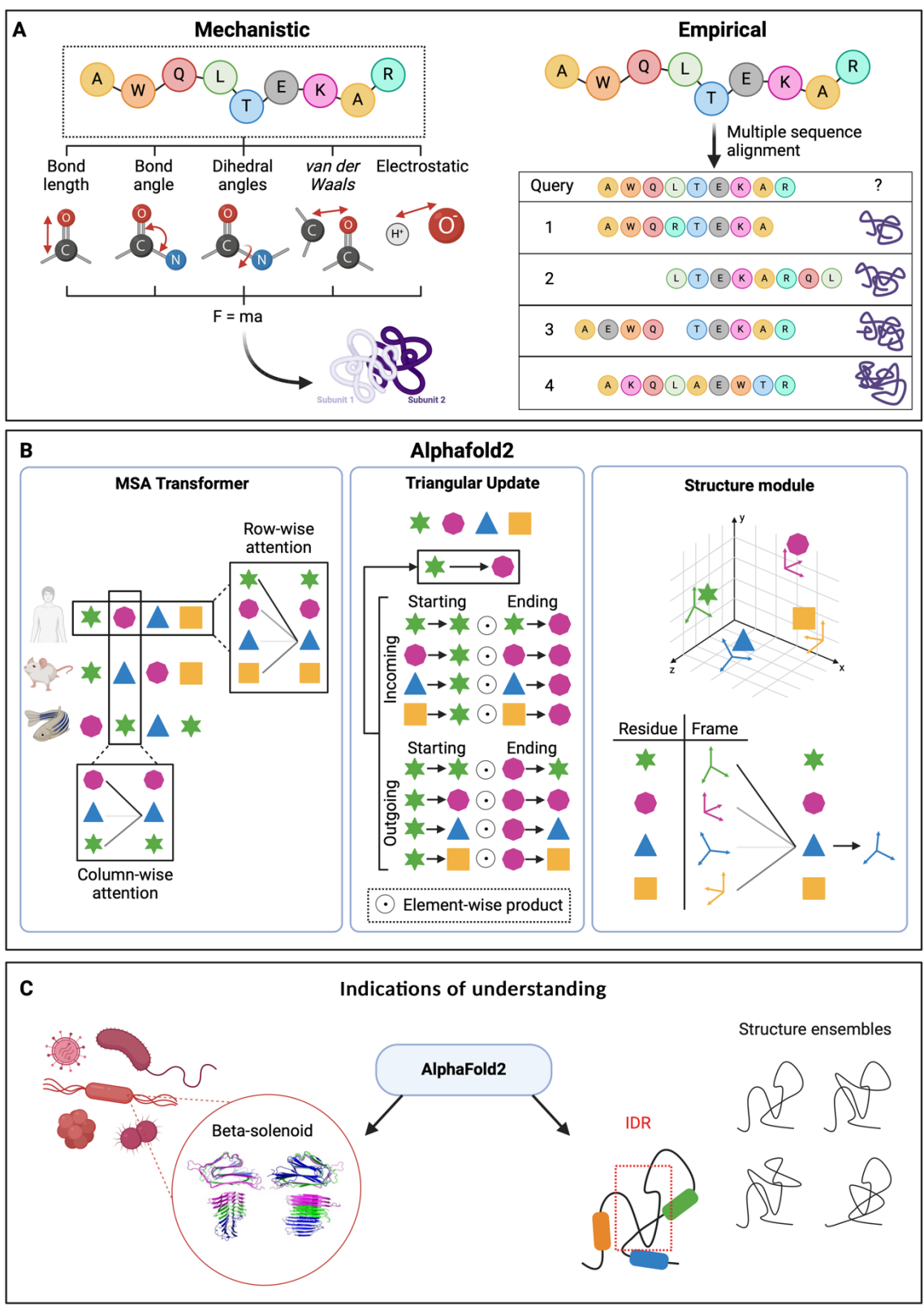

*Figure 3: AlphaFold2 is the most successful protein structure prediction model to date. A: We categorise traditional protein structure prediction task into two methodologies - mechanistic (template-free) and empirical (template-based). B: Three main components of the AlphaFold2 architecture - the MSA transformer, triangular update, and structure module. C: Two examples where AlphaFold2 has indicated understanding of biochemistry - the beta-solenoid structure and IDRs (intrinsically disordered regions).*

# Single-cell RNA sequencing

**Introduction**

A cell's expression of genes correlates with the proteins the cell produces and is therefore a key indication of the cell's biological activity. Until just over 10 years ago, scientific exploration of whole tissue, organ, and system function relied on gene expression data of pooled cells. More recently, the idea that every cell is unique in its gene expression, function, and fate has been supported by novel techniques that enable the study of genetic material within individual cells [79, 80]. These single cell genomics technologies - in particular single cell RNA-sequencing (scRNA-seq) - have become increasingly popular, enabling researchers to explore cellular systems at unprecedented resolution. More specifically, scRNA-seq has provided insight into the fundamental biological features of a wide variety of systems and processes, including biomarkers of specific cell states [81], interactions between cell populations [82], and patterns of cellular development [83]. The toolkit for analyzing these data sets relies heavily on machine learning, and though the development of ML approaches for single cell analysis is an extremely active area of research [32, 41, 47], the epistemological implications of using these methods for scRNA-seq analysis have not been explored. Unlike typical ML tasks like image recognition or text classification, ML is often applied in the context of single cell analysis as a tool to aid basic scientific research, generating or validating mechanistic hypotheses rather than providing definitive answers. It is therefore apparent that the interplay between ML models and human understanding has important implications for biological research in this domain. In this section we provide an overview of the key ML technologies used in scRNA-seq analysis and conceptualize them in terms of philosophical theories of understanding, providing a framework for interpreting how ML models are mediating understanding of cellular biology through the analysis of single cell data.

Before scRNA-seq, RNA transcriptomic measurements were collected using high-throughput bulk sequencing [84]. Bulk sequencing technologies measure the average gene expression within a mixed population of cells and allow researchers to compare whole samples obtained from different conditions. Bulk sequencing has been particularly useful in cancer biomarker identification and detection, where biopsy samples from healthy patients can be compared to patients with cancer to identify biomarkers of disease and subsequently used as a diagnostic or prognostic marker for new patients [85, 86]. Importantly, bulk sequencing does not allow for gene expression differences between cell types to be distinguished within a single sample, which is crucial for discovering patterns of cellular development or heterogeneity that influence disease or other biological processes. While scRNA-seq addresses these shortcomings, it introduces a new set of technical challenges that have motivated the widespread adoption of machine learning methods. While there is no formal consensus protocol for these analyses, there are various steps included in typical analysis pipelines, including dimensionality reduction and data visualisation, clustering, and trajectory analysis. Here we introduce approaches to each of these analytical tasks, exploring the ways in which they are advancing and limiting human understanding of biological phenomena.

**Dimensionality reduction**

In scRNA-seq datasets, each cell is typically represented as a vector in euclidean space with components representing the expression of each measured gene. Although single cell sequencing methods have lower sensitivity than their bulk counterparts, it is still common to obtain expression values for thousands of genes, producing an extremely high-dimensional representation for each cell. Furthermore, the typical number of samples collected in scRNA-seq (compared to bulk) is significantly larger, with bulk sequencing studies rarely collecting more than 100 samples but scRNA-seq experiments routinely measuring thousands of cellular transcriptomes. To tame this complexity, one of the first steps typically taken in an scRNA-seq analysis is to produce a representation of the data in 2 or 3 dimensions, while retaining the maximal amount of variation in the data [87] (Figure 4, panel A). These representations are then typically visualized in 2-D or 3-D plots for exploratory data analysis. The emphasis on information compression and qualitative interpretability of these dimensionality reduction and visualization methods aligns well with Wilkenfeld and De Regt's theories of understanding [16, 20], suggesting that these models are able to facilitate scientific understanding despite failing to serve as traditional explanatory models.

The longest standing approach to dimensionality reduction is PCA (Principal Component Analysis) [31], a linear dimensionality reduction method which identifies uncorrelated principal components that explain the largest possible amount of total variation in the data. Though classical PCA is often used in scRNA-seq anaylsis [87], a called ZIFA (zero inflated factor analysis) has also been developed, using a distributional inductive bias for modelling dropout events when performing dimensionality reduction [88]. These methods are commonly used to generate representations for downstream quantitative analyses like clustering and trajectory inference; methods which can still operate in high-dimensional spaces but benefit from having components with low amounts of variation removed from the original gene expression space (with thousands of dimensions). However, PCA's linearity constraint often prevents it from effectively representing scRNA-seq data in two or three dimensions for visualization purposes. Two other popular dimensionality reduction methods are t-SNE (t-distributed Stochastic Neighbour Embedding) [34] or UMAP (Uniform Manifold Approximation Projection) [35]. Unlike PCA, these algorithms are non-linear dimensionality reduction methods that rely on a graph representation of relationships between cells, produced by constructing edges between cells with similar gene expression profiles. t-SNE and UMAP process this graph and embed its nodes as points in a low-dimensional space, preserving local similarities between cells rather than global structure [28]. Importantly, since they use a graph representation that only provides relational information between neighboring cells, UMAP and t-SNE introduce a locality-related inductive bias. Specifically, non-linear dimensionality reduction approaches in general typically make the assumption that the relationships between cells that are close together in the space of gene expression values are more relevant, and therefore more important to preserve, than relationships between cells that are distant in gene expression space.

**Clustering**

Clustering algorithms are another area of focus in ML-driven single cell RNA-sequencing analysis (Figure 4, panel B), enabling researchers to categorise cells into distinct groups based on their RNA expression profiles [89]. By calculating similarity between cells and clustering similar cells together, all clustering algorithms fundamentally rely on a locality-related inductive bias; however, the ways in which similarity is defined vary between algorithms, with important consequences. The K-means algorithm [40], for instance, measures similarity with exact euclidean distance between the gene expression values of different cells, which provides a fine-grained view of cell similarity but is potentially susceptible to overfitting to technical artefacts. The popular Louvain [90] and Leiden [91] algorithms, on the other hand, operate on single cell graphs defined by constructing edges between cells based on their gene expression similarity. Since these graphs often remove exact distance information, they can be viewed as a means of information compression, which can facilitate understanding by backgrounding unimportant variation [16] but may unnecessarily discard biologically relevant features. Despite their mechanistic differences, all clustering approaches have the same output: a categorisation of individual cells. For the purposes of qualitative intelligibility [20], these categories pair naturally with dimensionality reduction techniques, allowing researchers to visualize relationships between clusters of cells and identify patterns in gene expression as they relate to groups of cells.

The clusters of cells obtained from these automated approaches can be subsequently identified with known cell types or subtypes using markers of those populations as references [92]. This manual process of labelling cells using domain-specific knowledge of the underlying biology not only improves the qualitative intelligibility of the results of a clustering by conceptualizing the clusters in terms of qualitative biological knowledge; it also serves to determine if the information compression performed by the model has retained biologically relevant information [16]. However, given that manual labelling is tedious and prone to bias, novel machine learning methods have also been developed to streamline and automate the clustering and cell-type identification tasks, including deep neural networks [93] and combinatorial platforms [94]. Though cell type identification analyses may lead to novel insights and facilitate qualitative interpretation [95], it is important to recognise that the categorisations proposed by a clustering will always incur a loss of information. The notion that cells should be categorized into distinct groups at all has recently been challenged by the identification of transitioning cells (cells that are transitioning from one type to another), perhaps implying that cells should be modelled on a continuous scale instead [96]. These ideas are the foundation of trajectory analysis.

**Trajectory inference**

Single cell RNA sequencing studies often limit the collection of cellular transcriptomic data to a single point in time, and with current technologies it is not yet possible to monitor the transcriptomes of individual cells over time. Thus, dynamical and temporal processes such as cell cycling rates, differentiation trajectories, and transient effects due to environmental changes are challenging to model. Cells captured at the same time may on average exhibit similar gene expression profiles that allow them to be grouped into one population, but in reality these cells may be in different states (e.g. different cell cycle stage or different stage in differentiation). Simpson's paradox describes this phenomenon, noting that a particular pattern in a larger population may disappear or reverse in particular subpopulations [97]. Simple clustering and annotation using the methods previously mentioned allow the user to select stages of dynamical processes of interest, but do not capture the full extent of cellular gene expression dynamics. Preparing cells for sequencing is already a laborious process, and it may not be feasible to manually separate cells from these states as different samples or synchronize them experimentally before collection.

Recent developments in scRNA-seq analysis have attempted to develop ML methods for resolving time-sensitive transcriptional differences without the need for time-series data [96, 98] (Figure 4, panel C). These methods aim to model dynamic temporal processes using trajectory inference (TI), which infers an ordering of cells along a continuous path that represents a time-dependent process based on their RNA expression profiles. Models for constructing this ordering typically rely on a locality-related inductive bias, assuming that cells with more similar gene expression profiles (according to some distance metric) are more likely to be adjacent along with inferred trajectory. These models have been applied to a wide range of dynamical biological systems including B cell development [99], blood cell differentiation [98] and neurogenesis [100].

Due to the high complexity of TI methods, qualitative interpretation of their outputs can be challenging, especially compared to standard discrete clustering methods. However, these challenges can be overcome to some extent using dimensionality reduction and visualization techniques that specifically preserve specific trajectory structures (for example, nonlinear branching) such as PHATE [101], and methods for differential gene expression that exploit continuous resolution such as tradeSeq [102]. Another open question is the extent to which the true chronology of gene expression events - the durations involved, measured in units of time - can be inferred from trajectories, since events such as transcriptional bursts can distort the relationship between time and distance along the trajectory [103]. We therefore note that both for the purposes of qualitative intelligibility and mechanistic accuracy it may be beneficial to study trajectories based solely on the relative ordering of cells rather than their exact positions along its path. Despite these caveats, TI clearly provides a valuable tool for understanding dynamic cellular processes from static scRNA-seq data sets.

**The future of ML in scRNA-seq**

Machine learning methods for scRNA-seq data analysis are increasingly being applied as a tool to understand cellular biological systems. By using ML tools for dimensionality reduction and visualization, clustering, and trajectory analysis, researchers are able to interpret and understand the high-dimensional gene expression better than was ever previously possible. These methods compress information into more compact representations and allow manipulation of the data into more qualitatively intelligible forms, facilitating scientific understanding without directly providing explanations of the underlying phenomena [16, 20]. Hence, ML has emerged in the scRNA-seq space not only as a predictive tool but as an effective way of illuminating the underlying mechanisms of cellular biology. However, challenges remain in translating the vast quantities of data produced by scRNA-seq experiments into useful insights. The simplification inherent in dimensionality reduction may lead to the loss of subtle but crucial features of cellular gene expression in the plots produced and interpreted by researchers. This is particularly applicable to non-linear methods like t-SNE and UMAP, which might sacrifice global structural preservation for enhanced local clustering through their locality-related inductive biases. Additionally, while clustering techniques facilitate qualitative interpretation of cell types and states, they can obscure the continuous and dynamic spectrum of cellular states and transitions. The emergence of trajectory inference methods attempts to address this limitation by providing a continuous framework to study cellular dynamics, yet the complexity of these methods may occasionally interfere with their intelligibility. Despite these limitations, the vast toolkit of ML methods for scRNA-seq data analysis clearly has great potential to advance scientific knowledge and allow researchers to obtain insights from prohibitively large and complex data sets.

As ML research in the context of scRNA-seq continues to advance, ML methods are becoming increasingly sophisticated and therefore difficult to use and interpret, creating a novel set of required skills for researchers who conduct scRNA-seq analysis. Researchers may have limited programming experience but can make use of online guides and tutorials that simplify and summarize the general workflow. Though these resources are important for improving accessibility, they mask the complexity of the underlying analysis and the importance of modelling choices made in the analytical workflow. Each ML model makes different assumptions under its particular inductive biases and it has been shown the choice of model and its hyperparameters can significantly influence the results of the analysis [104]. While it may sometimes be infeasible to reason theoretically about modelling choices, there exists a rich literature of benchmarking for different approaches to dimensionality reduction [32], clustering [105], and trajectory inference [96] under different conditions, which can serve as empirical justification for choosing one model over another in the context of a given analysis.

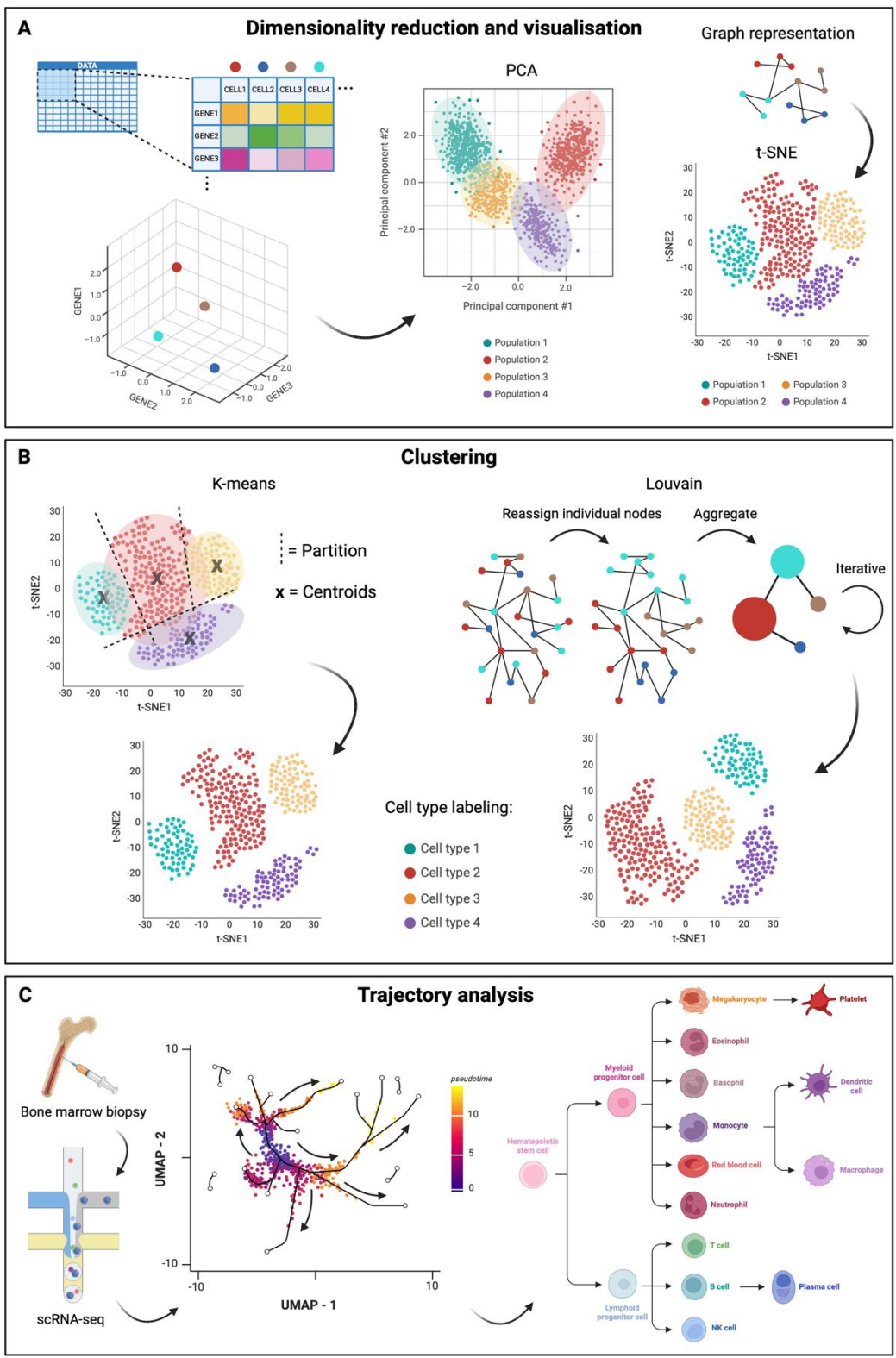

*Figure 4: Machine learning with single-cell RNA sequencing (scRNA-seq) data has provided insight into the features of a variety of biological systems and processes. We focus on three main steps in a typical scRNA-seq analysis pipeline. A: Dimensionality-reduction and visualisation, example approaches to which include PCA (principle component analysis) and t-SNE (t-distributed Stochastic Neighbour Embedding). B: Clustering, the process of identifies similar groups of cells. Two popular ML algorithms are K-means and Louvain. C: Trajectory inference, the process of reconstructing trajectories of cellular development without longitudinal measurements.*

## Conclusions

Biology in the 21st century has transformed into a data-driven science, relying on the acquisition and analysis of large data sets to probe the features of living systems. In recent years, machine learning approaches have been applied to a wide variety of biological problems with impressive results [3, 32, 37, 41, 47, 98]. Although ML-based modelling is being used across a wide range of scientific disciplines [1], life sciences in particular stands to benefit from the application of ML due to its need to model complex systems about which few general assumptions can be made. As a family of methods relying fundamentally on inductive reasoning, ML approaches are particularly suited for studying biological phenomena, which have historically resisted deductive explanation due to their multidimensional, conditional, and emergent properties. While these properties make deductive reasoning intractable in many cases, they also pose unique challenges in the application of ML models in biology, which have the potential to produce inaccurate, incomplete, or misleading outputs if they are not designed, interpreted, and applied appropriately. Furthermore, the application of ML methods for scientific inquiry ostensibly represents a significant shift in approach from the traditional scientific method, which has historically focused on explanatory unification of natural phenomena through laws and theories from which deductive reasoning can be applied. We propose that recent philosophical accounts of scientific understanding beyond explanation provide not only a theoretical characterisation of ML-mediated biological discovery, but also serve as a pragmatic framework for evaluating different technical aspects of ML-based modelling in biological research.

In this work we aimed to explore how ML is supporting efforts to understand biological systems through the lens of three conceptions of scientific understanding: Wilkenfeld's account [16], which highlights the importance of effectively compressing information, de Regt's account [20], which conceptualises scientific understanding as the capacity to reason qualitatively about target phenomena, and Dellsén's account [21], which emphasises the role of constructing an accurate model of the target system's dependencies. We provide an overview of these epistemological concepts, introduce key technical considerations for ML-based modelling in biology, and review two case studies to summarise how effective ML models of biological systems have been designed and how they have advanced understanding of their target phenomena. Specifically, we relate recent ML-driven advances in protein structure prediction and single-cell RNA-sequencing to the epistemological accounts of understanding as information compression, qualitative intelligibility, and dependency relation modelling, producing a general framework that can be used to guide how ML models of biological systems are designed, evaluated, and interpreted as tools for scientific understanding. Biological science will undoubtedly benefit from considering these philosophical foundations of understanding in the development and application of ML, producing ML systems that perform well at their task, provide robust guarantees for their behavior and function, and advance scientific knowledge of biological systems and phenomena.

# References

1. Wang, H., et al., *Scientific discovery in the age of artificial intelligence.* Nature, 2023. **620**(7972): p. 47-60.
2. Eraslan, G., Ž. Avsec, J. Gagneur, and F.J. Theis, *Deep learning: new computational modelling techniques for genomics.* Nature Reviews Genetics, 2019. **20**(7): p. 389-403.
3. Jumper, J., et al., *Highly accurate protein structure prediction with AlphaFold.* Nature, 2021. **596**(7873): p. 583-589.
4. Shen, J. and C.A. Nicolaou, *Molecular property prediction: recent trends in the era of artificial intelligence.* Drug Discov Today Technol, 2019. **32-33**: p. 29-36.
5. Friedman, M., *Explanation and Scientific Understanding.* The Journal of Philosophy, 1974. **71**(1): p. 5-19.
6. Hempel, C.G., *Aspects of scientific explanation.* Vol. 1. 1965: Free Press New York.
7. Railton, P., *A Deductive-Nomological Model of Probabilistic Explanation.* Philosophy of Science, 1978. **45**(2): p. 206-226.
8. Born, M. and E. Wolf, *Principles of Optics: Electromagnetic Theory of Propagation, Interference and Diffraction of Light.* 7 ed. 1999, Cambridge: Cambridge University Press.
9. Khalifa, K., *Understanding, Explanation, and Scientific Knowledge*, in *Understanding, Explanation, and Scientific Knowledge*, K. Khalifa, Editor. 2017, Cambridge University Press: Cambridge.
10. Strevens, M., *No understanding without explanation.* Studies in History and Philosophy of Science Part A, 2013. **44**(3): p. 510-515.
11. Lazebnik, Y., *Can a biologist fix a radio?--Or, what I learned while studying apoptosis.* Cancer Cell, 2002. **2**(3): p. 179-82.
12. Zhou, H.X., *Q&A: What is biophysics?* BMC Biol, 2011. **9**: p. 13.
13. Wolfram, S., *Undecidability and intractability in theoretical physics.* Physical Review Letters, 1985. **54**(8): p. 735-738.
14. Lundberg, S. and S.-I. Lee *A Unified Approach to Interpreting Model Predictions*. 2017. arXiv:1705.07874 DOI: 10.48550/arXiv.1705.07874.
15. Tulio Ribeiro, M., S. Singh, and C. Guestrin *"Why Should I Trust You?": Explaining the Predictions of Any Classifier*. 2016. arXiv:1602.04938 DOI: 10.48550/arXiv.1602.04938.
16. Wilkenfeld, D.A., *Understanding as compression.* Philosophical Studies, 2019. **176**(10): p. 2807-2831.
17. Feinstein, A., *A new basic theorem of information theory.* Transactions of the IRE Professional Group on Information Theory, 1954. **4**(4): p. 2-22.
18. Bronstein, M.M., J. Bruna, T. Cohen, and P. Veličković *Geometric Deep Learning: Grids, Groups, Graphs, Geodesics, and Gauges*. 2021. arXiv:2104.13478 DOI: 10.48550/arXiv.2104.13478.
19. Rauber, P.E., S.G. Fadel, A.X. Falcão, and A.C. Telea, *Visualizing the Hidden Activity of Artificial Neural Networks.* IEEE Transactions on Visualization and Computer Graphics, 2017. **23**(1): p. 101-110.
20. de Regt, H.W., *Understanding Scientific Understanding.* 2017: Oxford University Press.

21. Dellsén, F., *Beyond Explanation: Understanding as Dependency Modeling.* British Journal for the Philosophy of Science, 2018(4): p. 1261-1286.
22. Popper, K.R., *Conjectures and Refutations: The Growth of Scientific Knowledge*. Vol. 15. 1962: Routledge. 372.
23. Poincare, H., *Science and hypothesis*. Science and hypothesis., ed. G.B. Halsted. 1905, Oxford, England: Science Press. Pp. xxx+i, 196-Pp. xxx+i, 196.
24. Zhang, C., et al., *Understanding deep learning (still) requires rethinking generalization.* Commun. ACM, 2021. **64**(3): p. 107–115.
25. Ying, X., *An Overview of Overfitting and its Solutions.* Journal of Physics: Conference Series, 2019. **1168**(2): p. 022022.
26. Bender, A. and I. Cortés-Ciriano, *Artificial intelligence in drug discovery: what is realistic, what are illusions? Part 1: Ways to make an impact, and why we are not there yet.* Drug Discovery Today, 2021. **26**(2): p. 511-524.
27. Bender, A. and I. Cortes-Ciriano, *Artificial intelligence in drug discovery: what is realistic, what are illusions? Part 2: a discussion of chemical and biological data.* Drug Discovery Today, 2021. **26**(4): p. 1040-1052.
28. Karczewski, K.J. and M.P. Snyder, *Integrative omics for health and disease.* Nature Reviews Genetics, 2018. **19**(5): p. 299-310.
29. Bellman, R.E., *Adaptive Control Processes.* A Guided Tour. 1961: Princeton University Press.
30. Goodfellow, I., Y. Bengio, and A. Courville, *Representation Learning*, in *Deep learning*. 2016, MIT press Cambridge, MA, USA. p. 330-372.
31. Jolliffe, I.T. and J. Cadima, *Principal component analysis: a review and recent developments.* Philos Trans A Math Phys Eng Sci, 2016. **374**(2065): p. 20150202.
32. Xiang, R., et al., *A Comparison for Dimensionality Reduction Methods of Single-Cell RNA-seq Data.* Frontiers in Genetics, 2021. **12**.
33. Kramer, M.A., *Autoassociative neural networks.* Computers & Chemical Engineering, 1992. **16**(4): p. 313-328.
34. Hinton, G.E. and S. Roweis, *Stochastic Neighbor Embedding*, in *Advances in Neural Information Processing Systems*, S. Becker, S. Thrun, and K. Obermayer, Editors. 2002, MIT Press.
35. McInnes, L., J. Healy, and J. Melville *UMAP: Uniform Manifold Approximation and Projection for Dimension Reduction*. 2018. arXiv:1802.03426 DOI: 10.48550/arXiv.1802.03426.
36. LeCun, Y., Y. Bengio, and G. Hinton, *Deep learning.* Nature, 2015. **521**(7553): p. 436-444.
37. Stokes, J.M., et al., *A Deep Learning Approach to Antibiotic Discovery.* Cell, 2020. **180**(4): p. 688-702.e13.
38. Watson, J.L., et al., *De novo design of protein structure and function with RFdiffusion.* Nature, 2023. **620**(7976): p. 1089-1100.
39. Mitchell, T.M., *The need for biases in learning generalizations.* 1980.
40. MacQueen, J. *Some methods for classification and analysis of multivariate observations.* 1967.
41. Zhang, S., et al., *Review of single-cell RNA-seq data clustering for cell-type identification and characterization.* Rna, 2023. **29**(5): p. 517-530.
42. Langfelder, P. and S. Horvath, *WGCNA: an R package for weighted correlation network analysis.* BMC Bioinformatics, 2008. **9**(1): p. 559.
43. Ingraham, J., V. Garg, R. Barzilay, and T. Jaakkola, *Generative Models for Graph-Based Protein Design*, in *Advances in Neural Information Processing Systems*, H. Wallach, et al., Editors. 2019, Curran Associates, Inc.
44. Huang, S.-C., et al., *Self-supervised learning for medical image classification: a systematic review and implementation guidelines.* npj Digital Medicine, 2023. **6**(1): p. 74.

45. van de Schoot, R., et al., *Bayesian statistics and modelling.* Nature Reviews Methods Primers, 2021. **1**(1): p. 1.
46. Huelsenbeck, J.P., F. Ronquist, R. Nielsen, and J.P. Bollback, *Bayesian Inference of Phylogeny and Its Impact on Evolutionary Biology.* Science, 2001. **294**(5550): p. 2310-2314.
47. Nault, R., et al., *Benchmarking of a Bayesian single cell RNAseq differential gene expression test for dose–response study designs.* Nucleic Acids Research, 2022. **50**(8): p. e48-e48.
48. Love, M.I., W. Huber, and S. Anders, *Moderated estimation of fold change and dispersion for RNA-seq data with DESeq2.* Genome Biology, 2014. **15**(12): p. 550.
49. Veličković, P., *Everything is connected: Graph neural networks.* Current Opinion in Structural Biology, 2023. **79**: p. 102538.
50. Gilmer, J., et al. *Neural Message Passing for Quantum Chemistry.* 2017. arXiv:1704.01212 DOI: 10.48550/arXiv.1704.01212.
51. Vaswani, A., et al. *Attention Is All You Need.* 2017. arXiv:1706.03762 DOI: 10.48550/arXiv.1706.03762.
52. Lin, Z., et al., *Evolutionary-scale prediction of atomic-level protein structure with a language model.* Science, 2023. **379**(6637): p. 1123-1130.
53. Ruidong, W., et al., *High-resolution <em>de novo</em> structure prediction from primary sequence.* bioRxiv, 2022: p. 2022.07.21.500999.
54. Joshi, C., *Transformers are graph neural networks.* The Gradient, 2020.
55. Corso, G., et al. *DiffDock: Diffusion Steps, Twists, and Turns for Molecular Docking.* 2022. arXiv:2210.01776 DOI: 10.48550/arXiv.2210.01776.
56. Duval, A., et al., *A hitchhiker's guide to Geometric GNNs for 3D atomic systems.* 2023.
57. Anfinsen, C.B., E. Haber, M. Sela, and F.H. White, Jr., *The kinetics of formation of native ribonuclease during oxidation of the reduced polypeptide chain.* Proc Natl Acad Sci U S A, 1961. **47**(9): p. 1309-14.
58. Deng, H., Y. Jia, and Y. Zhang, *Protein structure prediction.* Int J Mod Phys B, 2018. **32**(18).
59. Dill, K.A., S.B. Ozkan, M.S. Shell, and T.R. Weikl, *The protein folding problem.* Annu Rev Biophys, 2008. **37**: p. 289-316.
60. Gandhi, J., et al., *Protein misfolding and aggregation in neurodegenerative diseases: a review of pathogeneses, novel detection strategies, and potential therapeutics.* Rev Neurosci, 2019. **30**(4): p. 339-358.
61. McCammon, J.A., B.R. Gelin, and M. Karplus, *Dynamics of folded proteins.* Nature, 1977. **267**(5612): p. 585-590.
62. Altschul, S.F., et al., *Gapped BLAST and PSI-BLAST: a new generation of protein database search programs.* Nucleic Acids Research, 1997. **25**(17): p. 3389-3402.
63. Eddy, S.R., *Profile hidden Markov models.* Bioinformatics, 1998. **14**(9): p. 755-63.
64. Baek, M., et al., *Accurate prediction of protein structures and interactions using a three-track neural network.* Science, 2021. **373**(6557): p. 871-876.
65. Kandathil, S.M., A.M. Lau, and D.T. Jones, *Machine learning methods for predicting protein structure from single sequences.* Current Opinion in Structural Biology, 2023. **81**: p. 102627.
66. Monastyrskyy, B., et al., *New encouraging developments in contact prediction: Assessment of the CASP11 results.* Proteins, 2016. **84 Suppl 1**(Suppl 1): p. 131-44.
67. Senior, A.W., et al., *Improved protein structure prediction using potentials from deep learning.* Nature, 2020. **577**(7792): p. 706-710.
68. Akdel, M., et al., *A structural biology community assessment of AlphaFold2 applications.* Nature Structural & Molecular Biology, 2022. **29**(11): p. 1056-1067.
69. Kajava, A.V. and A.C. Steven, *β-Rolls, β-Helices, and Other β-Solenoid Proteins*, in *Advances in Protein Chemistry*. 2006, Academic Press. p. 55-96.

70. Mesdaghi, S., R.M. Price, J. Madine, and D.J. Rigden, *Deep Learning-based structure modelling illuminates structure and function in uncharted regions of β-solenoid fold space.* Journal of Structural Biology, 2023. **215**(3): p. 108010.
71. Tunyasuvunakool, K., et al., *Highly accurate protein structure prediction for the human proteome.* Nature, 2021. **596**(7873): p. 590-596.
72. Necci, M., D. Piovesan, and S.C.E. Tosatto, *Critical assessment of protein intrinsic disorder prediction.* Nat Methods, 2021. **18**(5): p. 472-481.
73. Brotzakis, Z.F., Z. Shengyu, and V. Michele, *AlphaFold Prediction of Structural Ensembles of Disordered Proteins.* bioRxiv, 2023: p. 2023.01.19.524720.
74. Meller, A., S. Bhakat, S. Solieva, and G.R. Bowman, *Accelerating Cryptic Pocket Discovery Using AlphaFold.* Journal of Chemical Theory and Computation, 2023. **19**(14): p. 4355-4363.
75. Karplus, M. and J.A. McCammon, *Molecular dynamics simulations of biomolecules.* Nat Struct Biol, 2002. **9**(9): p. 646-52.
76. Echenique, P., *Introduction to protein folding for physicists.* Contemporary Physics, 2007. **48**(2): p. 81-108.
77. Duan, L., et al., *Accelerated Molecular Dynamics Simulation for Helical Proteins Folding in Explicit Water.* Front Chem, 2019. **7**: p. 540.
78. Fersht, A.R., *On the simulation of protein folding by short time scale molecular dynamics and distributed computing.* Proceedings of the National Academy of Sciences, 2002. **99**(22): p. 14122-14125.
79. Islam, S., et al., *Characterization of the single-cell transcriptional landscape by highly multiplex RNA-seq.* Genome Res, 2011. **21**(7): p. 1160-7.
80. Tang, F., et al., *mRNA-Seq whole-transcriptome analysis of a single cell.* Nature Methods, 2009. **6**(5): p. 377-382.
81. Kim, J., Z. Xu, and P.A. Marignani, *Single-cell RNA sequencing for the identification of early-stage lung cancer biomarkers from circulating blood.* npj Genomic Medicine, 2021. **6**(1): p. 87.
82. Efremova, M., M. Vento-Tormo, S.A. Teichmann, and R. Vento-Tormo, *CellPhoneDB: inferring cell–cell communication from combined expression of multi-subunit ligand–receptor complexes.* Nature Protocols, 2020. **15**(4): p. 1484-1506.
83. Lee, R.D., et al., *Single-cell analysis identifies dynamic gene expression networks that govern B cell development and transformation.* Nature Communications, 2021. **12**(1): p. 6843.
84. Emrich, S.J., W.B. Barbazuk, L. Li, and P.S. Schnable, *Gene discovery and annotation using LCM-454 transcriptome sequencing.* Genome Res, 2007. **17**(1): p. 69-73.
85. Chen, P.L., et al., *Analysis of Immune Signatures in Longitudinal Tumor Samples Yields Insight into Biomarkers of Response and Mechanisms of Resistance to Immune Checkpoint Blockade.* Cancer Discov, 2016. **6**(8): p. 827-37.
86. Han, L.O., et al., *Development and validation of an individualized diagnostic signature in thyroid cancer.* Cancer Med, 2018. **7**(4): p. 1135-1140.
87. Lun, A.T., D.J. McCarthy, and J.C. Marioni, *A step-by-step workflow for low-level analysis of single-cell RNA-seq data with Bioconductor.* F1000Res, 2016. **5**: p. 2122.
88. Pierson, E. and C. Yau, *ZIFA: Dimensionality reduction for zero-inflated single-cell gene expression analysis.* Genome Biology, 2015. **16**(1): p. 241.
89. Kiselev, V.Y., T.S. Andrews, and M. Hemberg, *Challenges in unsupervised clustering of single-cell RNA-seq data.* Nature Reviews Genetics, 2019. **20**(5): p. 273-282.
90. Blondel, V.D., J.-L. Guillaume, R. Lambiotte, and E. Lefebvre, *Fast unfolding of communities in large networks.* Journal of Statistical Mechanics: Theory and Experiment, 2008. **2008**: p. 10008.
91. Traag, V.A., L. Waltman, and N.J. van Eck, *From Louvain to Leiden: guaranteeing well-connected communities.* Scientific Reports, 2019. **9**(1): p. 5233.

92. Luecken, M.D. and F.J. Theis, *Current best practices in single-cell RNA-seq analysis: a tutorial.* Mol Syst Biol, 2019. **15**(6): p. e8746.
93. Yang, F., et al., *scBERT as a large-scale pretrained deep language model for cell type annotation of single-cell RNA-seq data.* Nature Machine Intelligence, 2022. **4**(10): p. 852-866.
94. Ianevski, A., A.K. Giri, and T. Aittokallio, *Fully-automated and ultra-fast cell-type identification using specific marker combinations from single-cell transcriptomic data.* Nature Communications, 2022. **13**(1): p. 1246.
95. Travaglini, K.J., et al., *A molecular cell atlas of the human lung from single-cell RNA sequencing.* Nature, 2020. **587**(7835): p. 619-625.
96. Saelens, W., R. Cannoodt, H. Todorov, and Y. Saeys, *A comparison of single-cell trajectory inference methods.* Nature Biotechnology, 2019. **37**(5): p. 547-554.
97. Simpson, E.H., *The Interpretation of Interaction in Contingency Tables.* Journal of the Royal Statistical Society. Series B (Methodological), 1951. **13**(2): p. 238-241.
98. Qiu, X., et al., *Reversed graph embedding resolves complex single-cell trajectories.* Nature Methods, 2017. **14**(10): p. 979-982.
99. Bendall, S.C., et al., *Single-cell trajectory detection uncovers progression and regulatory coordination in human B cell development.* Cell, 2014. **157**(3): p. 714-25.
100. Shin, J., et al., *Single-Cell RNA-Seq with Waterfall Reveals Molecular Cascades underlying Adult Neurogenesis.* Cell Stem Cell, 2015. **17**(3): p. 360-72.
101. Moon, K.R., et al., *Visualizing structure and transitions in high-dimensional biological data.* Nature Biotechnology, 2019. **37**(12): p. 1482-1492.
102. Van den Berge, K., et al., *Trajectory-based differential expression analysis for single-cell sequencing data.* Nature Communications, 2020. **11**(1): p. 1201.
103. Haque, A., J. Engel, S.A. Teichmann, and T. Lönnberg, *A practical guide to single-cell RNA-sequencing for biomedical research and clinical applications.* Genome Medicine, 2017. **9**(1): p. 75.
104. Raimundo, F., C. Vallot, and J.-P. Vert, *Tuning parameters of dimensionality reduction methods for single-cell RNA-seq analysis.* Genome Biology, 2020. **21**(1): p. 212.
105. Yu, L., Y. Cao, J.Y.H. Yang, and P. Yang, *Benchmarking clustering algorithms on estimating the number of cell types from single-cell RNA-sequencing data.* Genome Biology, 2022. **23**(1): p. 49.